\documentclass[letterpaper,10pt]{article}
\usepackage{tabularx} 
\usepackage{amsmath}  
\usepackage[margin=1in]{geometry} 
\usepackage{cite} 
\usepackage[final]{hyperref} 
\hypersetup{
	colorlinks=true,       
	linkcolor=blue,        
	citecolor=blue,        
	filecolor=magenta,     
	urlcolor=blue         
}

\usepackage{blindtext}
\usepackage{natbib}
\usepackage[utf8]{inputenc} 
\usepackage[T1]{fontenc}    
\usepackage{hyperref}       
\usepackage{url}            
\usepackage{booktabs}       
\usepackage{amsfonts}       
\usepackage{nicefrac}       
\usepackage{microtype}      
\usepackage{xcolor}         
\usepackage{mathrsfs}
\usepackage{dsfont}
\usepackage{amssymb}  
\usepackage{makecell}
\usepackage{amsmath}
\usepackage{multirow}
\usepackage{booktabs}
\usepackage{algorithm}
\usepackage{algpseudocode}
\usepackage{amsthm}
\theoremstyle{definition}

\usepackage{xspace}
\usepackage{longtable}
\usepackage{wrapfig}
\usepackage{enumitem}
\usepackage{etoc}
\setlist{leftmargin=5mm}
\usepackage[export]{adjustbox}

\usepackage{todonotes}
\usepackage{xcolor}

\usepackage{amsmath}
\usepackage{mdframed}
\usepackage{graphicx}
\usepackage{subcaption}
\usepackage{pifont}
\usepackage{times}

\title{Learning from ``Silly'' Questions Improves Large Language Models,\\ But Only Slightly}

\author{
    Tingyuan Zhu\thanks{Institute of Science Tokyo. Email: \texttt{zhu.t.ac@m.titech.ac.jp}} \and
    Shudong Liu\thanks{University of Macau. Email: \texttt{nlp2ct.shudong@gmail.com}} \and
    Yidong Wang\thanks{Peking University} \and
    Derek F. Wong\footnotemark[2] \and
    Han Yu\thanks{Nanyang Technological University} \and
    Takahiro Shinozaki\footnotemark[1] \and
    Jindong Wang\thanks{William \& Mary. Email: \texttt{jwang80@wm.edu}}
}

\date{}

\newmdenv[
  linecolor=black,
  linewidth=1pt,
  roundcorner=5pt,
  innertopmargin=1.5\baselineskip,
  innerbottommargin=1.5\baselineskip,
  innerrightmargin=1em,
  innerleftmargin=1em,
  backgroundcolor=gray!5,
  font=\ttfamily 
]{custombox}

\begin{document}

\maketitle

\begin{abstract}
Constructing high-quality Supervised Fine-Tuning (SFT) datasets is critical for the training of large language models (LLMs).
Recent studies have shown that using data from a specific source, Ruozhiba\footnote{\url{https://huggingface.co/datasets/m-a-p/COIG-CQIA/viewer/ruozhiba}.}, a Chinese website where users ask ``silly'' questions to better understand certain topics, can lead to better fine-tuning performance. This paper aims to explore some hidden factors: the potential interpretations of its success and a large-scale evaluation of the performance. First, we leverage GPT-4 to analyze the successful cases of Ruozhiba questions from the perspective of education, psychology, and cognitive science, deriving a set of explanatory rules. Then, we construct fine-tuning datasets by applying these rules to the MMLU training set. Surprisingly, our results indicate that rules can significantly improve model performance in \emph{certain} tasks, while potentially diminishing performance on others. For example, SFT data generated following the ``Counterintuitive
Thinking" rule can achieve approximately a 5\% improvement on the ``Global Facts" task, whereas the ``Blurring the Conceptual Boundaries" rule leads to a performance drop of 6.14\% on the ``Econometrics" task. In addition, for specific tasks, different rules tend to have a consistent impact on model performance. This suggests that the differences between the extracted rules are not as significant, and the effectiveness of the rules is relatively consistent across tasks. Our research highlights the importance of considering task diversity and rule applicability when constructing SFT datasets to achieve more comprehensive performance improvements.
\end{abstract}

%

\section{Introduction}

Large language models (LLMs), pre-trained on vast amounts of data, have garnered significant attention for their ability to address a wide range of tasks~\citep{llama3, gpt4, gemini, qwencoder, qwenmath, Mixtral, liu-etal-2024-llms-learn-uncertainty, zhou2024your, zhao2023survey}. The pretrain-finetune paradigm has emerged as the cornerstone of LLMs' remarkable success. Through pre-training, LLMs acquire extensive knowledge about the world, while fine-tuning aligns them with specific human instructions, enabling them to generate high-quality responses and excel on domain-specific datasets. Compared to the resource-intensive process of pre-training, supervised fine-tuning is a cost-effective approach that focuses on fine-tuning LLMs using small but high-quality datasets. Consequently, the quality of SFT datasets plays a pivotal role in determining the performance of fine-tuned models.

Numerous studies~\citep{Zhou2023LIMALI, Liu2023WhatMG, mekala-etal-2024-smaller, xia2024less, li-etal-2024-one, Chen2023AlpaGasusTA, Lu2023InsTagIT, cao2023instruction, wei2023instructiongpt} have emphasized the importance of the quality of SFT datasets. The scarcity of high-quality datasets necessitates careful data selection and processing when constructing SFT datasets. Recently, \citet{bai2024coig} demonstrated that SFT data collected from certain sources can significantly improve fine-tuning results. Notably, they highlighted the significant contribution of the data from ``Ruozhiba'', a Chinese online platform where people ask ``silly'' questions like ``Since 70\% of the human body is water, does that mean 7 out of every 10 people are just water disguised as humans?'' to better understand certain knowledge.
Ruozhiba contains implicit satire, critical content, and elements that are somewhat aggressive and offensive; such data are generally not considered suitable for SFT. However, \citet{bai2024coig} argue that an important reason for the good results with the Ruozhiba is that its dialogues often exhibit characteristics of humor, absurdity, linguistic traps, and abstraction that can enhance the model's reasoning capabilities.

\begin{figure}[t!]
    \centering
    \includegraphics[width=0.9\textwidth]{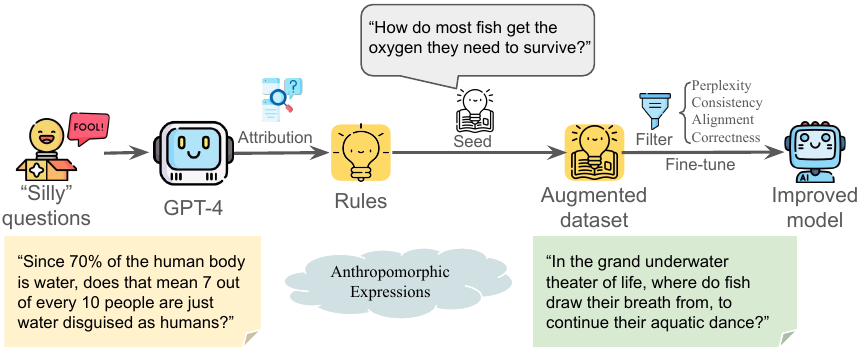}
    \caption{Overview of our augmentation pipeline.}
    \label{fig:overview}
\end{figure}

This work attempts to present a holistic understanding of data augmentation in Ruozhiba style from two aspects: the potential interpretation of its effectiveness and a large-scale evaluation.
Specifically, from the perspective of education, psychology and social learning, we used GPT-4 to extract features from the Ruozhiba dataset and defined these features as augmentation rules. We then employed GPT-4o to augment seed data sampled from MMLU, generating datasets that conform to different rules. Subsequently, we fine-tuned the LLM using both the seed datasets and the generated datasets and compared the performance changes to assess the impact of the augmentation.

Our experimental results show that fine-tuning LLMs with datasets generated using “silly” rules can achieve up to approximately a 0.54\% overall performance improvement on the MMLU test set. However, compared to directly fine-tuning with the seed dataset, this approach does not yield any further overall performance gains. Our more fine-grained analysis reveals that datasets generated using different rules have \emph{varying} impacts on the performance of the SFT model across different subjects and tasks. At the subject level, the extracted rules tend to degrade the performance of the SFT model on “STEM” subject, whereas some rules lead to slight improvements in  “Humanities” subject. Furthermore, our task-level analysis shows that for specific tasks, datasets generated with different rules tend to have consistent impacts. Based on our analysis of a 13,000-sample dataset, $94.74\%$ of tasks exhibited at least $50\%$ consistency in the performance impact percentages of different rules, and $26.32\%$ of tasks showed $100\%$ consistency. Additionally, our subject-level analysis showed that all subjects demonstrated over $60\%$ consistency in the performance impact percentages of different rules.


In summary, our contributions are as follows:  

\begin{enumerate}  
    \item \emph{Understanding:} We provide a holistic understanding of the Ruozhiba-style dataset from the perspective of education, psychology, and cognitive science. Our study generated a set of useful rules that can potentially impact the LLM fine-tuning performance.
    \item \emph{Augmentation:} We generate diverse datasets using the rules extracted from Ruozhiba to fine-tune LLMs. Our large-scale evaluation demonstrated varying improvements on different tasks.

    \item \emph{Selection:} We explore various data filtering and mixing strategies and find that SFT datasets enhanced with single-rule augmentation are more effective in improving LLM performance compared to mixed datasets. 

\end{enumerate}

\section{Related work}

\textbf{Post-training data synthesis:} The scarcity of high-quality SFT data has spurred research into synthesizing data using existing LLMs. For example, \citet{vicuna2023} collected user-shared ChatGPT conversations to build the SFT dataset Sharegpt, which was used for fine-tuning. After fine-tuning, the model achieved 90\% of the performance quality compared to closed-source models on the MT-Bench benchmark \citep{zheng2023judging}. The process of constructing the ShareGPT dataset can be viewed as a form of black-box distillation from closed-source models. More creative SFT data synthesis strategies leverage LLM capabilities to heuristically generate SFT data. The self-instruct method \citep{self-instruct} used bootstrapping to generate new instructions and corresponding input-output samples from a limited set of manually written task seeds, achieving a 33\% absolute improvement in GPT-3's performance on the SUPERNI benchmark~\citep{superbench} compared to the original GPT-3 model.

Building on this idea, \citet{Bommasani2021OnTO} used the Self-Instruct framework to create a 52k-sample Alpaca SFT dataset, fine-tuning Llama to achieve performance comparable to text-davinci-003 on the self-instruct validation set. WizardLM \citep{Xu2023WizardLMEL} employed GPT as an instruction evolver, creating SFT data with varying complexity through depth and breadth evolution from seed data. Furthermore, \citet{zeng-automatic} optimized the WizardLM by automating the analysis, summarization, and optimization of instruction dataset evolution strategies, thereby enhancing LLM performance across diverse tasks. These studies highlight the potential of innovative data synthesis strategies in advancing LLM capabilities.

\textbf{Supervised fine-tuning:} also known as instruction tuning, is a crucial component of the post-training paradigm for LLMs. SFT involves fine-tuning LLMs, which have undergone next-token prediction, on high-quality datasets composed of (instruction, output) pairs. This process helps bridge the gap between model predictions and the knowledge or preferences of human users. For instance, InstructGPT~\citep{instructGPT} demonstrated that fine-tuning LLMs with high-quality SFT datasets annotated by humans significantly enhances the model's ability to follow user instructions. Similarly, the study \citet{chung2024scaling} highlighted that fine-tuning language models on a set of instruction-formulated datasets can improve model performance and generalization capability on unseen tasks.

Given the substantial resource demands of fully fine-tuning large models, parameter-efficient fine-tuning (PEFT) methods have gained prominence. Techniques like LoRA~\citep{lora} and its variants~\citep{longlora, relora, mora} update the dense neural network layers of LLMs with plug-in low-rank matrices, achieving comparable performance to full fine-tuning while keeping the LLM parameters frozen. This approach has become a mainstream method for PEFT in large models.

\textbf{Post-training data selection:}
Compared to the vast amounts of data used during pre-training, post-training emphasizes achieving significant improvements with a smaller quantity of high-quality data. LIMA \citep{Zhou2023LIMALI} demonstrated that using just 1,000 carefully constructed SFT data points can significantly boost the performance of mainstream open-source LLMs like LLama on downstream tasks. Another study \citep{Liu2023WhatMG} investigated SFT data selection from the perspectives of complexity, quality, and diversity, showing that models fine-tuned with 6,000 SFT training samples perform comparably to baseline models trained with tens of thousands of samples. Furthermore, \citet{mekala-etal-2024-smaller} advocated for the ability of current language models to autonomously select high-quality training data, designing a perplexity (ppl)-based data selection method that effectively trains models to perform on par with those trained on complete datasets.

COIG-CQIA \citep{bai2024coig} collected data from various sources to construct an SFT dataset. They found that fine-tuning LLMs with the Ruozhiba subset significantly outperformed other data sources on the BELLE-EVAL benchmark \citep{belle2023exploring}. The authors attributed this to the unique characteristics of the Ruozhiba dataset, which includes cognitive and linguistic traps, jokes, riddles, and artistic and abstract rhetorical techniques, thereby enhancing the LLM's language understanding and multi-hop logical reasoning capabilities. These studies collectively underscore the importance of data selection in SFT.

\section{Methodology}

In this section, we will articulate the approach to understanding the Ruozhiba questions, augmenting training data, and model fine-tuning.
\figurename~\ref{fig:overview} illustrates our pipeline.

\subsection{Understanding ``Silly'' Questions}
It requires significant effort to characterize the factors attributing to the success of the ``silly questions'' in Ruozhiba.
Formally speaking, given an existing Ruozhiba dataset (e.g., data from COIG-CQIA~\citep{bai2024coig}) $D_{base}=\{\mathrm{ins}_i, \mathrm{rep}_i\}_{i=1}^n$, where $\mathrm{ins}_i$ and $\mathrm{rep}_i$ denote instruction and response, respectively, our objective is to learn a set of rules $R = \{r_j\}_{j=1}^T$ to interpret why $D_{base}$ helps fine-tuning.

However, given the complicated nature of human language, especially the ``nonsense'' contained in the existing Ruozhiba dataset, it is extremely challenging to manually analyze and summarize the hidden factors in the dataset.
In this paper, we take advantage of LLMs to facilitate our analysis.
Specifically, we prompt GPT-4 from the perspective of psychology, education, sociology, and cognitive science to extract and summarize explanatory rule from $D_{base}$.
These rules encapsulate the inherent logic and structure of tasks in the dataset, which not only aid in better understanding the dataset itself but also provide guidance and a foundation for constructing new high-quality datasets. Through this approach, we hope that the generated datasets will achieve diversity and applicability across various tasks, thereby enhancing the performance of LLMs in a wide range of tasks.

\subsection{Data Augmentation}

Given a seed dataset $D_{seed}=\{d_1, d_2, \ldots, d_N\}$, where 
$d_i$ consists of a pair $\{\mathrm{ins}_i, \mathrm{rep}_i\}$, we utilize the GPT-4 as Instruction Rewriter to transform each $\mathrm{ins}_i$ by referencing a rule $r_t \in R$. This transformation results in a new instruction $\mathrm{ins}_{i, r_t}$ that adheres to the rule $r_t$, forming new SFT pairs $d_{i, r_t} = \{\text{ins}_{i, r_t}, \text{rep}_i\}$. Consequently, we obtain a dataset $D_{r_t}$ following rule $r_t$.
Additionally, we include $\mathrm{res}_i$ in the prompt to ensure that the rewritten $\mathrm{ins}_{i, r_t}$ can still be accurately responded to with $\text{rep}_i$.
The prompts are in Appendix~\ref{sec-app-prompt-rewriter}.

Recognizing that the Instruction Rewriter might not fully comprehend the rule $r_t$, we manually select several data points from $D_{base}$ as in-context examples (see Appendix~\ref{sec-app-prompt} for details). These examples serve to guide the Instruction Rewriter in better accomplishing the task.

To explore the varying effects of different rules across subjects or tasks, we designed experiments involving data filtering and mixing. For parallel data samples \(d_{i, r_1}, d_{i, r_2}, \dots, d_{i, r_t}\), we generated a mixed sample \(d_{i, r_{\text{mix}}}\) using the following equation:
$$
d_{i, r_{\text{mix}}} = \text{Filter}(d_{i, r_1}, d_{i, r_2}, \dots, d_{i, r_t}),
$$
where $\text{Filter}$ denotes the filtering operation for data selection.
Specifically, we performed the following strategies:
\begin{enumerate}
    \item \textbf{Perplexity-Based Selection}.
    We proposed two strategies to select samples based on their perplexity (PPL), either the samples with the highest or the lowest perplexity:
    $$
    d_{i, r_{\text{mix}}} = d_{i, r_k}, \quad \text{where } k = \arg\max_{k} \; \text{PPL}(d_{i, r_k}),
    $$
    $$
    d_{i, r_{\text{mix}}} = d_{i, r_k}, \quad \text{where } k = \arg\min_{k} \; \text{PPL}(d_{i, r_k}).
    $$
    When implementing the perplexity-based selection, we also considered whether to include seed data, resulting in four configurations.
    \item \textbf{Judge Model Scoring-Based Selection}.
    We utilized a Judge model to evaluate augmented samples across three dimensions: \textit{Consistency}, \textit{Correctness}, and \textit{Alignment}. For each dimension, we selected the sample with the highest score:
    $$
    d_{i, r_{\text{mix}}} = d_{i, r_k}, \quad \text{where } k = \arg\max_{k} \; \text{Judge}(d_{i, r_k}).
    $$
\end{enumerate}

\begin{figure}[t!]
    \centering
    \includegraphics[width=.8\textwidth]{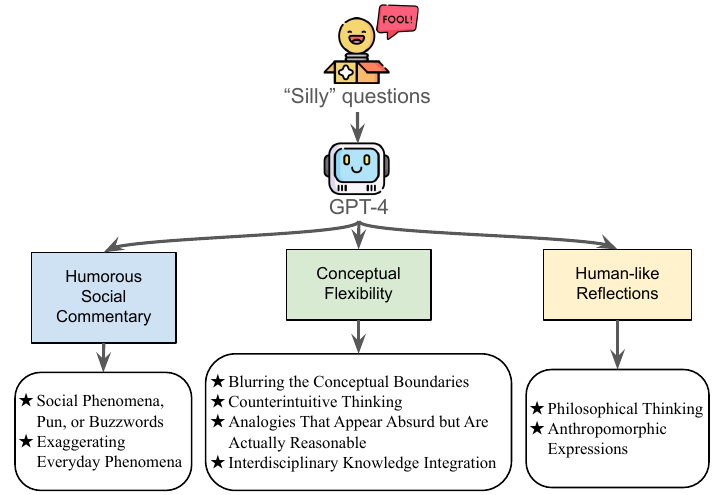}
    \caption{The rules extracted from the Ruozhiba dataset.}
    \label{fig:rule_extraction}
\end{figure}

\begin{figure*}[htbp]
    \centering
    \includegraphics[width=1\textwidth]{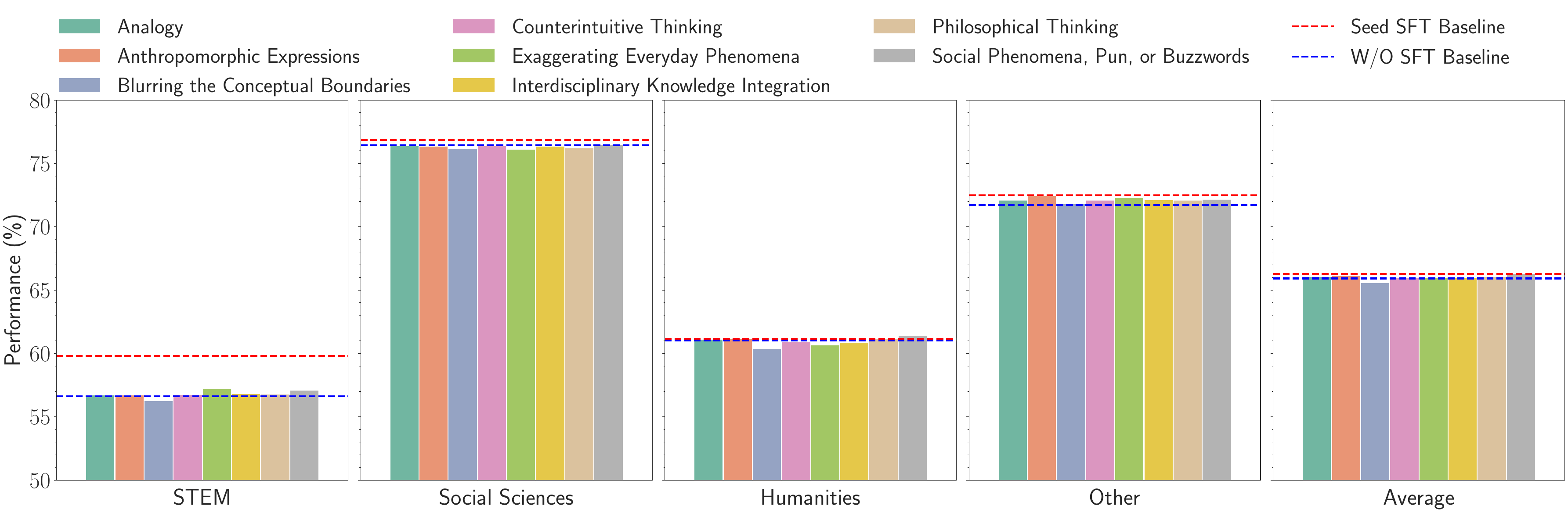}
    \caption{We conduct SFT on the language model LLM using the seed dataset \( D_{\text{seed}} \) and eight rule-generated datasets. Subject-level accuracy is evaluated on the MMLU test set.}
    \label{fig:subject_analysis}
\end{figure*}

These strategies resulted in seven distinct mixing configurations. Due to API cost constraints, we applied all seven configurations to a subset containing 4K samples. For the larger 13K dataset, we only employed perplexity-based filtering strategies.

\section{Experiments}

\subsection{Dataset}

We utilize the Ruozhiba dataset as \(D_{base}\) and extract eight rules, as shown in \figurename~\ref{fig:rule_extraction}. For detailed information about the Ruozhiba dataset, please refer to Appendix~\ref{sec-app-rule-data}. To validate the effectiveness of our rules across a broad spectrum of domain knowledge, we select the MMLU train set to sample \(D_{seed}\). MMLU is a large-scale, multi-task, multiple-choice question-answering dataset encompassing 57 tasks across natural sciences, humanities, and social sciences. Utilizing MMLU as the seed dataset enables us to assess the impact of extracted rules on diverse tasks, such as knowledge-based QA and logical reasoning, since the influence of rules is likely to vary across different task types. This dataset allows us to rigorously evaluate the adaptability of our rules across a wide array of knowledge domains.

The MMLU dataset provides a training set containing approximately 90K samples from datasets such as ARC~\citep{arc}, MC-TEST~\citep{mctest}, OBQA~\citep{obqa}, and RACE~\citep{race}. We performed stratified sampling on the training set to ensure a balanced representation of tasks. Consequently, we sampled approximately 13K instances to form our seed dataset, \(D_{seed}\), with the task distribution illustrated in \figurename~\ref{fig:sampled_distribution} of Appendix~\ref{sec-app-exp-setup}.

Using the rules extracted in the last section: \(R = \{r_1, r_2, \ldots, r_8\}\), we employed GPT-4 as the instruction rewriter to process \(D_{seed}\), resulting in eight distinct datasets: \(D_{r1}, D_{r2}, \ldots, D_{r8}\). These datasets, along with \(D_{seed}\), constitute our experimental dataset. To construct these datasets, we made 13K × 8 = 104K API requests.

\begin{figure}[htbp]
    \centering
    \includegraphics[width=0.6\columnwidth]{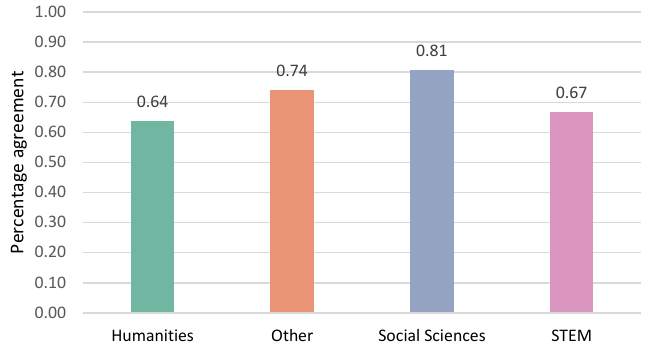}
    \caption{The subject-level percentage agreement for datasets generated by the eight rules, considering whether SFT results improved, declined, or remained unchanged relative to the model fine-tuned with $D_{seed}$.}
    \label{fig:agreement_ratio}
\end{figure}

\subsection{General Results}
We fine-tune Meta-Llama-3-8B-Instruct using the LoRA setup. For detailed experimental configurations, please refer to Appendix~\ref{sec-app-exp-setup}. We evaluate the performance on the original dataset \(D_{\text{seed}}\) as well as on datasets synthesized according to eight different rules. The results are illustrated in \figurename~\ref{fig:subject_analysis}. It is evident that the effectiveness of the rules varies across different subjects. Overall, datasets generated by these rules, after fine-tuning, underperformed in “STEM” subjects compared to directly fine-tuning with \(D_{\text{seed}}\). Whereas in the “Humanities” and “Other” subjects, the rule-generated datasets showed superior performance compared to \(D_{\text{seed}}\). On the entire MMLU test set (i.e., the Average columns), the dataset generated by the rule for “Social Phenomena, Pun, and Buzzwords” slightly outperformed \(D_{\text{seed}}\) (66.28 vs 66.27), whereas other rule-generated datasets did not achieve higher scores.

Given that MMLU's four subjects comprise 57 tasks, we conducted a more fine-grained task-level analysis, detailed in the Appendix~\ref{sec-app-exp-results}. We categorized the performance of different SFT datasets compared with \(D_{\text{seed}}\) into \emph{improved}, \emph{declined}, and \emph{unchanged} categories, and calculated the consistency percentage of these performance impacts. Our analysis revealed that for the same task, datasets generated by different rules tend to produce consistent changes in performance. As illustrated in \figurename~\ref{fig:agreement_ratio}, our subject-level analysis shows that the consistency percentage of performance impacts from different rules exceeded 60\% across all subjects. Furthermore, in our task-level analysis, 94.73\% of the tasks demonstrated a consistency percentage over 50\%, and for 26.32\% of the tasks, the consistency reached 100\%.


We conduct experiments on data filtering and mixing strategies. Due to API costs, we perform the PPL-based filtering and mixing experiment on the 13K dataset and apply all data mixing strategies to a 4K subset. The experimental results are provided in the Appendix~\ref{sec-app-exp-results}. Our findings indicate that none of the mixing strategies surpass the optimal results achieved by using a single rule-based augmentation method.

Overall, using Ruozhiba's rules for data augmentation on the MMLU dataset has not resulted in significant improvements. We believe the reasons are mainly as follows:

\begin{enumerate}
    \item As shown in the task-level analysis results in the Appendix~\ref{sec-app-exp-results}, the tasks in MMLU vary greatly, encompassing different domains of knowledge (such as natural sciences and social sciences) and different types of tasks (such as knowledge-based questions and logical reasoning). Improvements in some tasks may be offset by declines in others.
    
    \item As shown in the in-context examples in Appendix~\ref{sec-app-prompt-rewriter}, Ruozhiba's data combines human knowledge with linguistic ingenuity, and not all texts can be rewritten to achieve such clever expression. In the samples generated using the rules provided in Appendix~\ref{sec-app-example}, it is evident that GPT-4's rewrites primarily focus on stylistic changes. However, it remains challenging to mimic the nuanced expressions found in the in-context examples.
    
    \item \citet{bai2024coig} used the BELLE-EVAL benchmark, where GPT serves as the judge. Within the COIG-CQIA dataset, Ruozhiba is unique for employing GPT in data filtering and response annotation. This is significant because LLM-as-a-Judge can exhibit \textit{self-enhancement bias}~\citep{zheng2023judging}, potentially favoring its own outputs and inflating Ruozhiba's performance. To mitigate this, we calculated the accuracy of multiple-choice questions and had GPT-4 rewrite the questions while keeping the options and responses unchanged, effectively removing this bias. This might explain why the Ruozhiba dataset didn't show significant improvement in our experiments.
\end{enumerate}

\section{Conclusion}
In summary, our study focuses on the Ruozhiba data source to explore the nuanced effects of specific data characteristics on the SFT of LLMs. Our findings underscore the complexity and variability in model performance across different subjects and tasks, influenced by distinct data generation rules. Notably, while rules extracted from Ruozhiba tend to diminish performance in STEM-related tasks compared to seed datasets, they offer modest improvements in areas such as Philosophical Thinking and Social Phenomena, Pun, or Buzzwords. Furthermore, our analysis reveals a consistency in the impact of different rules on the performance of models fine-tuned for specific subjects or tasks, suggesting that the choice of generation rules may be less critical than their application to appropriate tasks. This insight is crucial for guiding the development of high-quality SFT datasets, highlighting the importance of tailoring data characteristics to specific domains and tasks to optimize LLM performance. 

\bibliographystyle{plainnat}
\bibliography{refs}

\onecolumn
\appendix

\setcounter{section}{0}
\renewcommand{\thesection}{\Alph{section}}

\begin{center}
    {\LARGE \textbf{Appendix}}
\end{center}

By extracting these rules, we aim to guide the Instruction Rewriter in generating high-quality SFT datasets. We provide in-context examples with the expectation that the Instruction Rewriter will follow the identified rules. This approach ultimately aims to enhance the performance of LLMs across various tasks by ensuring the generated datasets are aligned with these interdisciplinary and nuanced rules.

\section{Details of Rules}
\label{sec-app-rule}

In this section, we articulate the rules in \figurename~\ref{fig:rule_extraction} extracted from the Ruozhiba dataset.

\subsection{Humorous Social Commentary}
\subsubsection{Social Phenomena, Pun, or Buzzwords:}
   The use of puns and buzzwords adds humor and relatability, making social commentaries engaging and thought-provoking. By incorporating witty language and familiar phrases, the discussion becomes more accessible and entertaining, capturing readers' attention and encouraging reflection on underlying issues in a light-hearted manner. This method enhances the impact of the commentary, making complex topics more digestible and stimulating meaningful conversations.

\subsubsection{Exaggerating Everyday Phenomena:}
   Some instances in the dataset use exaggeration and literal interpretations to create absurd and humorous scenarios. This approach challenges readers to think critically about the logical extensions of everyday phenomena and the potential pitfalls of taking things too literally or out of context. It entertains while encouraging deeper reflection on rules, interpretations, and statistical measures often taken for granted.

\subsection{Conceptual Flexibility}
\subsubsection{Blurring the Conceptual Boundaries:}
   This rule involves challenging implicit assumptions by applying ideas from one scenario to another. It reveals nuanced, context-dependent truths that highlight the limitations of rigid, one-size-fits-all thinking. By examining assumptions through contrasting contexts, it encourages a more flexible and adaptable mindset, appreciating the complexity of interactions in our physical, social, and conceptual environments.

\subsubsection{Counterintuitive Thinking:}
   Counterintuitive thinking serves as a lens to view familiar situations in unexpected ways, often by purposefully misapplying assumptions or questioning norms. This approach exposes flaws in our thinking patterns and challenges the logic we take for granted.

\subsubsection{Analogies That Appear Absurd but Are Actually Reasonable:}
   Analogies play a crucial role in illuminating the interconnectedness of actions, the limitations of human perception, and the complexity of decision-making. They make abstract concepts tangible, highlight hidden connections, provoke reflection, and facilitate effective communication, thereby enhancing our understanding of life's complexities.
   
\subsubsection{Interdisciplinary Knowledge Integration:}
   This rule highlights the use of a cross-disciplinary approach to reframe everyday perceptions by combining scientific or logical methods. It aims to examine human behavior and social norms, challenging everyday assumptions with grounded, factual approaches to often subjective or socially driven phenomena.

\subsection{Human-like Reflections}
\subsubsection{Philosophical Thinking:}
   These contents encourage the grasp of nuanced and abstract concepts, enhancing the ability to interpret complex human thought patterns and language. By processing abstract thinking patterns, the model learns to generate responses that go beyond rigid, literal interpretations. This is crucial for engaging with open-ended, speculative questions, helping align more closely with human thought processes in creative, philosophical, or hypothetical discussions.

\subsubsection{Anthropomorphic Expressions:}
   This pattern involves exploring complex concepts by attributing human-like qualities to non-human elements, inviting a playful yet thought-provoking reexamination of everyday assumptions. These expressions question societal norms and values by comparing physical or economic realities to personal experiences, creating a lens of irony and curiosity. Through humor and paradox, they highlight the human tendency to accept certain systems without questioning their deeper implications, suggesting a reevaluation of taken-for-granted aspects.

\section{Rule Dataset}
\label{sec-app-rule-data}

Our research focuses on exploring whether the rules embedded in the Ruozhiba data from COIG-CQIA can enhance the performance of SFT models. For this purpose, we selected the Ruozhiba dataset from COIG-CQIA for rule extraction. This dataset originates from Baidu Tieba, the largest Chinese forum, and its content often uses metaphors, puns, and language traps to satirize social phenomena or create humor. The authors of COIG-CQIA, after scraping the data, conducted rigorous filtering to remove non-instructive and harmful content, ultimately selecting 240 questions. Subsequently, they utilized GPT-4 to annotate these data and continuously prompted GPT-4 with human input to optimize the responses, thereby avoiding language traps and obtaining high-quality 240 pairs of $\{\mathrm{ins}, \mathrm{rep}\}$. These optimized data not only retain the complex linguistic structures and humorous elements of the original data but also ensure their effectiveness and safety in training SFT models. In this way, they aimed to construct a high-quality supervised fine-tuning dataset to enhance the performance of large language models in multi-task environments.

\section{Rule Generation}

We utilized the Ruozhiba dataset that consists of 240 data points as in-context examples for GPT-4. The goal was to extract relevant rules based on interdisciplinary knowledge from psychology, sociology, education, and cognitive science. The complete prompt used for this extraction process is detailed in the Appendix~\ref{sec-app-prompt}. Through this process, we extracted eight rules as shown in \figurename~\ref{fig:rule_extraction}.

\section{Details of Prompts}
\label{sec-app-prompt}

In this section, we further present all the prompts we used in our study. The base prompt is modified based on the prompt used by \citet{Xu2023WizardLMEL}.

\subsection{Prompt for the Instruction Rewriter}
\label{sec-app-prompt-rewriter}

\subsubsection{Base Prompt:}
\begin{custombox}
\sloppy
I want you to act as a Prompt Rewriter. Your objective is to rewrite a given prompt into a more complex version to make those famous AI systems (e.g., chatgpt and GPT4) a bit harder to handle. But the rewritten prompt must be reasonable and must be understood and responded by humans. Your rewriting cannot omit the non-text parts such as the table and code in the original prompt. Also, please do not omit the input in the original prompt. You SHOULD complicate the given prompt using the following method: [Rule-specific prompt] You should try your best NOT to make the rewritten prompt become verbose; the rewritten prompt can only add 10 to 20 words into the original prompt. You should ensure that any names of people or entities are retained. Certain terms like 'original prompt', 'rewritten prompt', and related phrases are NOT allowed to appear in the rewritten prompt. It is also NOT allowed to include reference choices content in the rewritten prompt.
\end{custombox}

\subsubsection{Social Phenomena, Pun, or Buzzwords:}
\begin{custombox}
\sloppy
Please rewrite \#The Given Prompt\# using puns based on social phenomena, pun or buzzwords, while preserving the original intent as much as possible. Note that I prefer you to refer to the relevant knowledge rather than directly using specific terms. Regarding using puns based on social phenomena or buzzwords, here are some examples for your reference:

1. Experts suggest reducing the use of electronic devices, but isn’t there no material in the world that doesn’t contain electronics?

2. I bought a set of toys with the 26 letters of the alphabet, but I only received 23 of them. When I contacted the seller about it, they said I didn't purchase the DLC.

3. The saying goes, \"Only with pressure comes motivation.\" So, can atmospheric pressure also provide motivation?

\#The Given Prompt\#: [The Given Prompt]

\#Reference Choices\#:[Reference Choices]

\#Reference Answer\#: [Reference Answer]

\#Rewritten Prompt\#:
\end{custombox}
\subsubsection{Exaggerating Everyday Phenomena:}
\begin{custombox}
\sloppy
Please rewrite \#The Given Prompt\# by exaggerating everyday phenomena, while preserving the original intent as much as possible. Note that I prefer you to refer to the relevant knowledge rather than directly using specific terms. Regarding exaggerating everyday phenomena, here are some examples for your reference:

1. If concentrated sulfuric acid contains 2\% water, does drinking 50 cups of sulfuric acid mean I would have consumed a cup of pure water?

2. Is it a violation to run after drinking all the water during a swimming competition?

3. If a couple keeps getting married and divorced repeatedly, does that count as contributing to the marriage rate?

\#The Given Prompt\#: [The Given Prompt]

\#Reference Choices\#:[Reference Choices]

\#Reference Answer\#: [Reference Answer]

\#Rewritten Prompt\#:
\end{custombox}
\subsubsection{Blurring the Conceptual Boundaries:}
\begin{custombox}
\sloppy
Please rewrite \#The Given Prompt\# by blurring the conceptual boundaries, while preserving the original intent as much as possible. Note that I prefer you to refer to the relevant knowledge rather than directly using specific terms. Regarding blurring the conceptual boundaries, here are some examples for your reference:

1. After intense exercise, one should avoid drinking ice-cold water and taking cold showers. So what should one do if they're engaging in vigorous activity in the water?

2. Since a rusty knife can cause tetanus when used to cut someone, why didn't ancient people simply use rusty weapons?

3. Everyone works to make money, so who is losing money?

\#The Given Prompt\#: [The Given Prompt]

\#Reference Choices\#:[Reference Choices]

\#Reference Answer\#: [Reference Answer]

\#Rewritten Prompt\#:
\end{custombox}
\subsubsection{Counterintuitive Thinking:}
\begin{custombox}
\sloppy
Please rewrite \#The Given Prompt\# using counterintuitive thinking, while preserving the original intent as much as possible. Note that I prefer you to refer to the relevant knowledge rather than directly using specific terms. Regarding using counterintuitive thinking, here are some examples for your reference:

1. Since the prison is full of criminals, why don't the police go into the prison to arrest people?

2. A person should look for cars when crossing the street, but what should they do if there are no cars on the road?

3. If a surgery has a success rate of only 50\%, then doing it twice would result in a success rate of just 25\%. So, if we only perform half of the surgery, wouldn't that give us a 100\% success rate?

\#The Given Prompt\#: [The Given Prompt]

\#Reference Choices\#:[Reference Choices]

\#Reference Answer\#: [Reference Answer]

\#Rewritten Prompt\#:
\end{custombox}
\subsubsection{Analogies That Appear Absurd but Are Actually Reasonable:}
\begin{custombox}
\sloppy
Please   \#The Given Prompt\# using an analogy that appears absurd but is actually reasonable, while preserving the original intent as much as possible. Note that I prefer you to refer to the relevant knowledge rather than directly using specific terms. Regarding using an analogy that appears absurd but is actually reasonable, here are some examples for your reference:

1. On a rainy day, I stepped into a puddle and accidentally shattered the sky.

2. We are all blind people in life, groping the elephant named the world.

3. If I shatter the nightmare, am I the destroyer of dreams or the guardian of happiness?

\#The Given Prompt\#: [The Given Prompt]

\#Reference Choices\#:[Reference Choices]

\#Reference Answer\#: [Reference Answer]

\#Rewritten Prompt\#:
\end{custombox}
\subsubsection{Interdisciplinary Knowledge Integration:}
\begin{custombox}
\sloppy
Please rewrite \#The Given Prompt\# by integrating interdisciplinary knowledge, while preserving the original intent as much as possible. Note that I prefer you to refer to the relevant knowledge rather than directly using specific terms. Regarding integrating interdisciplinary knowledge, here are some examples for your reference:

1. It's clearly food-grade stainless steel—so why is it still so hard to swallow?

2. The baby raised by wolves can understand wolf language, so will a baby raised by robots know C language?

3. Many girls tend to exaggerate their age, so why not just look at their annual rings?

\#The Given Prompt\#: [The Given Prompt]

\#Reference Choices\#:[Reference Choices]

\#Reference Answer\#: [Reference Answer]

\#Rewritten Prompt\#:
\end{custombox}
\subsubsection{Philosophical Thinking:}
\begin{custombox}
\sloppy
Please rewrite \#The Given Prompt\# by incorporating philosophical thinking, while preserving the original intent as much as possible. Note that I prefer you to refer to the relevant knowledge rather than directly using specific terms. Regarding incorporating philosophical thinking, here are some examples for your reference:

1. The alarm clock shatters the dream; does it also shatter the life of another world?

2. The cry of a newborn baby—is it a joyful expression of life, or a fear of the world?

3. The meaning of life is to find happiness, so why does everyone seem to live in a sea of suffering?

\#The Given Prompt\#: [The Given Prompt]

\#Reference Choices\#:[Reference Choices]

\#Reference Answer\#: [Reference Answer]

\#Rewritten Prompt\#:
\end{custombox}
\subsubsection{Anthropomorphic Expressions:}
\begin{custombox}
\sloppy
Please rewrite \#The Given Prompt\# using anthropomorphic expressions, while preserving the original intent as much as possible. Note that I prefer you to refer to the relevant knowledge rather than directly using specific terms. Regarding using anthropomorphic expressions, here are some examples for your reference:

1. Since 70\% of the human body is water, does that mean 7 out of every 10 people are just water disguised as humans?

2. When a person is away from money, they are useless; when money is away from a person, it is just a piece of paper. So why do people work hard to earn money instead of money working hard to find people?

3. Why do we take medicine when we are sick, but the world seems to resort to sacrificing people when it is unwell? Can a person heal through medication, while the world can only be cured by taking lives?

\#The Given Prompt\#: [The Given Prompt]

\#Reference Choices\#:[Reference Choices]

\#Reference Answer\#: [Reference Answer]

\#Rewritten Prompt\#:
\end{custombox}
\subsection{Prompt for the Judge}
\begin{custombox}
\sloppy
I want you to act as a Data Synthesis Evaluator.
Your objective is to evaluate synthesized data against given standards to ensure quality and relevance.
You will be provided with an original question, the answer, the synthesis rule, and the synthesized question.
Your evaluation should be based on the following criteria, each rated on a scale of 1-10:

1. Consistency: Does the synthesized question align with the original question?

2. Correctness: Is the synthesized question accurate and free of errors?

3. Alignment: Does the synthesized question adhere to the specified synthesis rules?

Original Question: [Original Question]

Answer: [Answer]

Synthesis Rule: [Synthesis Rule]

Synthesized Question: [Synthesized Question]

Please provide your evaluation in the following structured format:

---

Consistency: [Score 1-10]

Correctness: [Score 1-10]

Alignment:  \quad [Score 1-10]

---
\end{custombox}

\section{Detailed Experiments}
\label{sec-app-exp}
In this section, we present our experimental details and the task distribution of the datasets.




\subsection{Experimental Details}
\label{sec-app-exp-setup}

In the supervised fine-tuning of the Meta-Llama-3-8B-Instruct model~\citep{llama3}, we configured hyperparameters to optimize the balance between model performance and computational efficiency. The per-device training batch size was set to 1, complemented by 8 gradient accumulation steps. We employed a learning rate of \(5 \times 10^{-6}\), utilizing a cosine learning rate scheduler to ensure gradual decay of the learning rate. A warmup ratio of 0.1 was implemented to stabilize initial training phases. The model was trained for 6 epochs, a duration chosen to optimize convergence while minimizing the risk of overfitting. We evaluated the model after SFT on MMLU using the 5-shot setting.

\begin{figure*}[!htp]
    \centering
    \includegraphics[width=\textwidth]{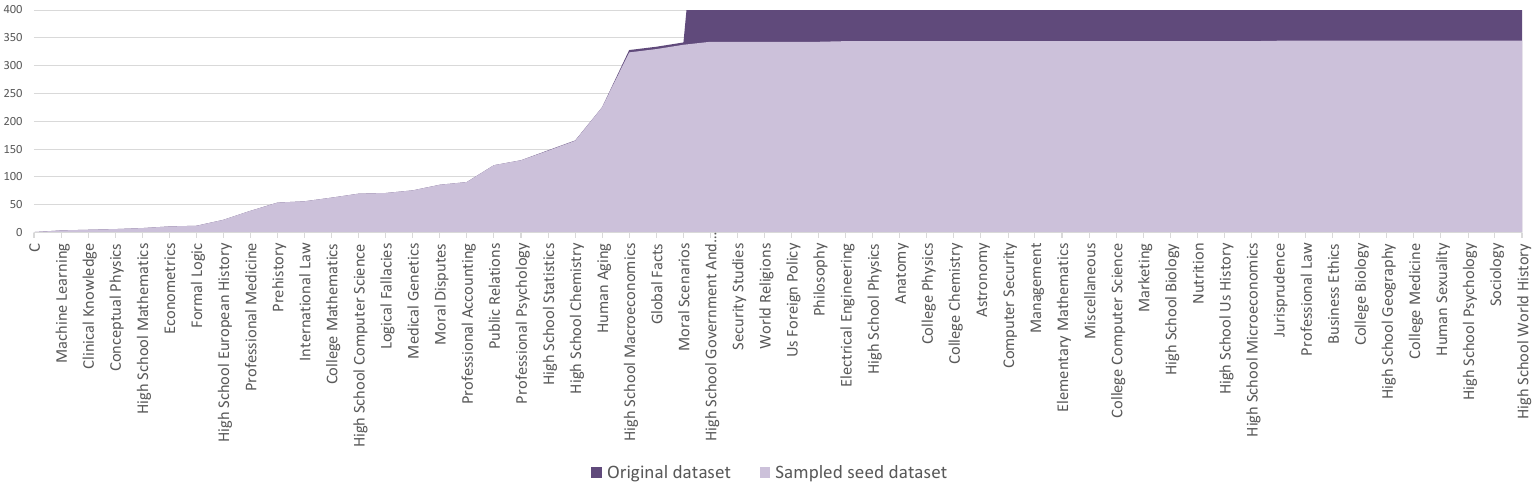}
    \caption{We sampled the MMLU training set to include all 57 tasks, ensuring diverse task coverage for the experiment.}
    \label{fig:sampled_distribution}
\end{figure*}

\newpage

\section{Examples of Generated Data}
\label{sec-app-example}

In this section, we present a seed example from the \(D_{seed}\) dataset, along with eight parallel examples generated by the Instruction Rewriter, each rewritten according to one of the eight specified rules.


\begin{custombox}
\sloppy
\quad
\textbf{Seed sample:}

\quad

How do most fish get the oxygen they need to survive?

\quad

\textbf{Social Phenomena, Pun, or Buzzwords:}

\quad

In the vast aquatic landscape, where fish navigate their liquid homes, how do these finned denizens engage in the biohack of extracting that life-sustaining gas, famously known as oxygen, from their watery world?

\quad

\textbf{Exaggerating Everyday Phenomena:}

\quad

By what miraculous process do most aquatic creatures extract the vital essence of life, allowing them to thrive beneath the surface of their watery realm?

\quad

\textbf{Blurring the Conceptual Boundaries:}

\quad

In the aquatic realm, the majority of these finned creatures acquire the essential life-sustaining element required for their survival. Could you explain the intricate process by which they manage to extract this invisible resource, necessary for their existence, from the environment that envelops them?

\quad

\textbf{Counterintuitive Thinking:}

\quad

In an environment where creatures spend their lives submerged in water and not on land, how do these beings extract the essential life-sustaining element needed from a medium that does not provide it in its gaseous form?

\quad

\textbf{Analogies That Appear Absurd but Are Actually Reasonable:}

\quad

By navigating the ocean's symphony, how do fish conduct their own orchestra to extract the elusive breath of life they require to survive?

\quad

\textbf{Interdisciplinary Knowledge Integration:}

\quad

In the context of aquatic life, how do most fish acquire the essential element for cellular respiration, considering the principles of fluid dynamics and the solubility of gases in liquid environments?

\quad

\textbf{Philosophical Thinking:}

\quad

What is the mechanism by which most aquatic beings procure the vital essence of oxygen necessary for their continued existence? Consider the interplay between their physical structures and the surrounding life-giving medium.

\quad

\textbf{Anthropomorphic Expressions:}

\quad

In the grand underwater theater of life, where do fish draw their breath from, to continue their aquatic dance?
\end{custombox}

\subsection{Detailed Results}
\label{sec-app-exp-results}

In this section, we present the task-level experimental results, including the accuracy of the SFT model on the MMLU test set under different experimental settings, as well as the task-level percentage agreement.

\begin{figure*}[!hbp]
    \centering
    \includegraphics[width=\textwidth]{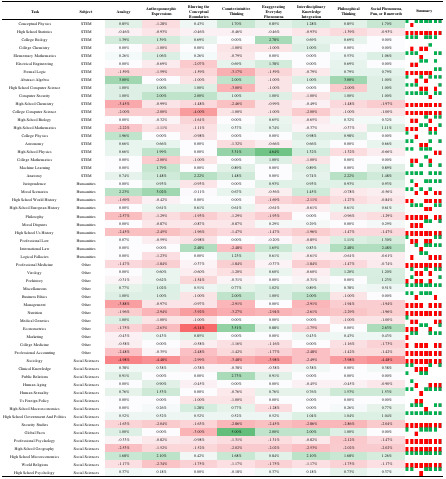}
    \caption{The task-level accuracy changes on eight rule-generated datasets (13K samples) compared to fine-tuning with $D_{seed}$.}
    \label{fig:overall_summarize_13000_1}
\end{figure*}

\begin{figure*}[t]
    \centering
    \includegraphics[width=\textwidth]{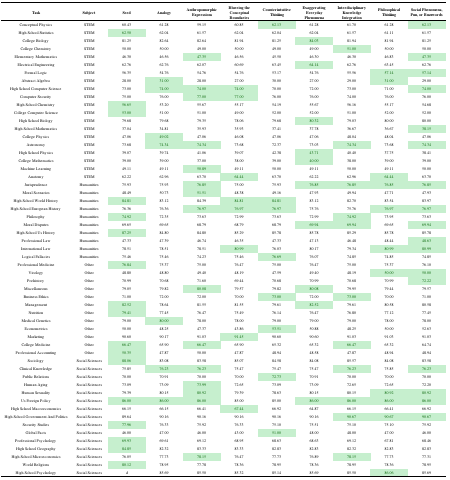}
    \caption{The task-level accuracy on eight rule-generated datasets (13K samples) compared to fine-tuning with $D_{seed}$.}
    \label{fig:category_analysis}
\end{figure*}

\begin{figure*}[t]
    \centering
    \includegraphics[width=\textwidth]{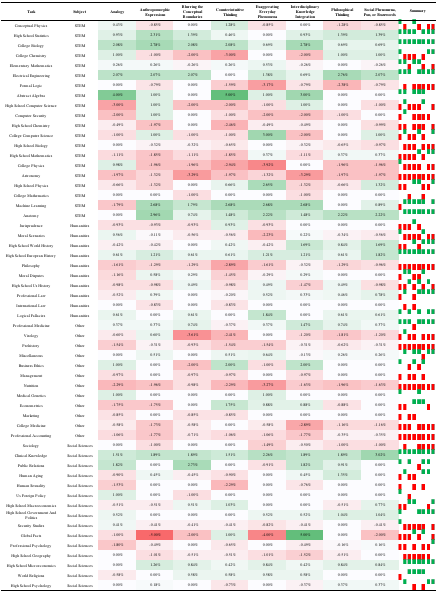}
    \caption{The task-level accuracy changes on eight rule-generated datasets (4K samples) compared to fine-tuning with $D_{seed}$.}
    \label{fig:overall_summarize_13000_1}
\end{figure*}

\begin{figure*}[t]
    \centering
    \includegraphics[width=\textwidth]{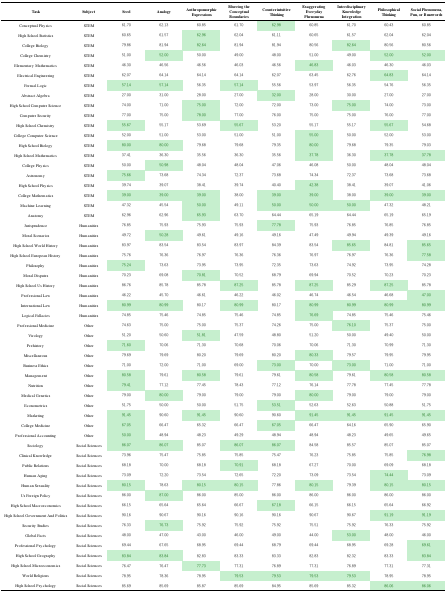}
    \caption{The task-level accuracy on eight rule-generated datasets (4K samples) compared to fine-tuning with $D_{seed}$.}
    \label{fig:category_analysis}
\end{figure*}

\begin{figure*}[t]
    \centering
    \includegraphics[width=\textwidth]{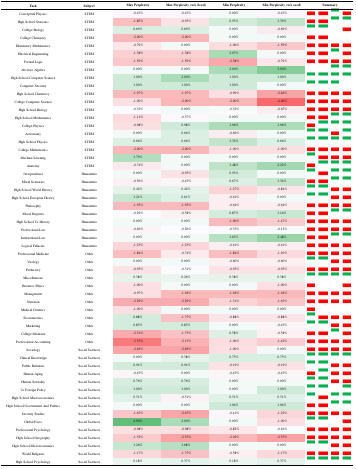}
    \caption{The task-level accuracy changes on four filtering and mixing strategies-generated datasets (13K samples) compared to fine-tuning with \( D_{\text{seed}} \).}
    \label{fig:overall_summarize_13000_1}
\end{figure*}

\begin{figure*}[t]
    \centering
    \includegraphics[width=\textwidth]{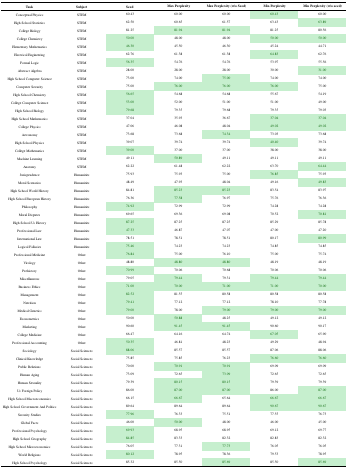}
    \caption{The task-level accuracy changes on four filtering and mixing strategies-generated datasets (13K samples) compared to fine-tuning with \( D_{\text{seed}} \).}
    \label{fig:category_analysis}
\end{figure*}

\begin{figure*}[t]
    \centering
    \includegraphics[width=\textwidth]{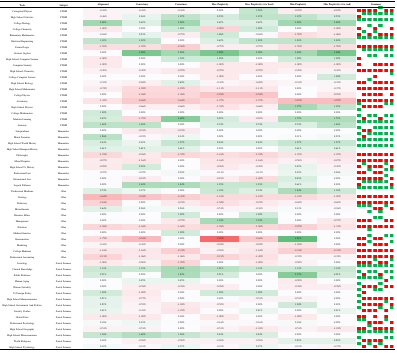}
    \caption{The task-level accuracy changes on seven filtering and mixing strategies-generated datasets (4K samples) compared to fine-tuning with \( D_{\text{seed}} \).}
    \label{fig:overall_summarize_13000_1}
\end{figure*}

\begin{figure*}[t]
    \centering
    \includegraphics[width=\textwidth]{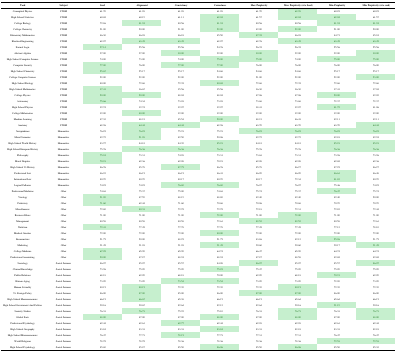}
    \caption{The task-level accuracy on seven filtering and mixing strategies-generated datasets (4K samples) compared to fine-tuning with \( D_{\text{seed}} \).}
    \label{fig:category_analysis}
\end{figure*}



\begin{figure}[htbp]
    \centering
    \begin{subfigure}[t]{0.45\textwidth}
        \centering
        \includegraphics[width=\textwidth]{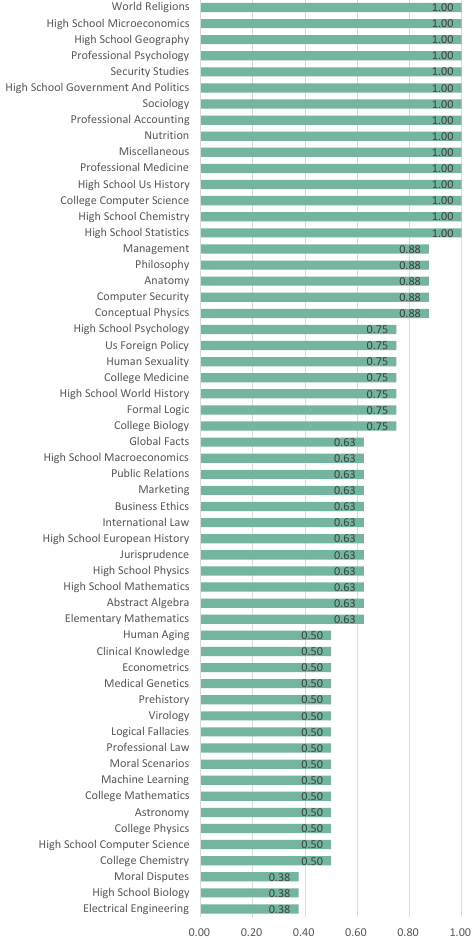}
        \caption{13K samples}
        \label{fig:13000_ratio}
    \end{subfigure}
    ~
    \begin{subfigure}[t]{0.45\textwidth}
        \centering
        \includegraphics[width=\textwidth]{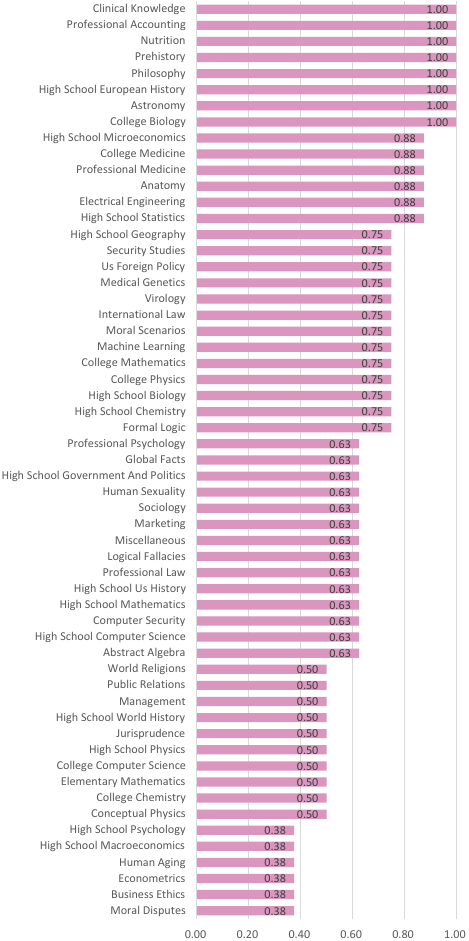}
        \caption{4K samples}
        \label{fig:4000_ratio}
    \end{subfigure}
    \caption{Comparison of task-level percentage agreement for datasets generated by the eight rules, considering whether SFT results improved, declined, or remained unchanged relative to the model fine-tuned with $D_{seed}$.}
    \label{fig:agreement_ratio}
\end{figure}

\end{document}